\documentclass[a4paper,11pt]{article}
\usepackage{pos}
\usepackage{braket}
\usepackage{graphicx}
\usepackage{dcolumn}
\usepackage{bm}
\usepackage{xcolor}
\usepackage{mathtools}
\usepackage{soul}
\setstcolor{red}
\usepackage{subcaption}
\usepackage{tikz}
\usepackage{booktabs}
\usepackage{multirow}
\usepackage{adjustbox}
\usepackage{comment}
\usepackage{physics}

\DeclareMathOperator{\sgn}{sgn}

\title{Quantum Qualifiers for Neural Network Model Selection in Hadronic Physics}

\author*{Brandon B. Le}
\author{D. Keller}

\affiliation{Department of Physics, University of Virginia,\\
  Charlottesville, Virginia 22904, USA}

\emailAdd{sxh3qf@virginia.edu}
\emailAdd{dmk9m@virginia.edu}

\abstract{As quantum machine-learning architectures mature, a central challenge is no longer their construction, but identifying the regimes in which they offer practical advantages over classical approaches. In this work, we introduce a framework for addressing this question in data-driven hadronic physics problems by developing diagnostic tools—centered on a quantitative quantum qualifier—that guide model selection between classical and quantum deep neural networks based on intrinsic properties of the data. Using controlled classification and regression studies, we show how relative model performance follows systematic trends in complexity, noise, and dimensionality, and how these trends can be distilled into a predictive criterion. We then demonstrate the utility of this approach through an application to Compton form factor extraction from deeply virtual Compton scattering, where the quantum qualifier identifies kinematic regimes favorable to quantum models. Together, these results establish a principled framework for deploying quantum machine-learning tools in precision hadronic physics.}

\FullConference{
The 26th international symposium on spin physics (SPIN2025)\\
21–26 September, 2025\\
Qingdao (Tsingtao), Shandong Province, China\\}

\begin{document}
\maketitle

\section{Introduction}

Quantum machine learning has opened new possibilities for data analysis in strongly correlated and information-rich physical systems, motivating growing interest in whether quantum models can offer practical advantages over classical approaches. In particular, quantum deep neural networks (QDNNs) employ parameterized quantum circuits as trainable models that operate in a Hilbert space and naturally encode complex correlations \cite{Beer2020,Biamonte2017}. While the extent of quantum advantage in near-term applications remains an open question, QDNNs provide a flexible framework for exploring how quantum architectures may outperform classical deep neural networks (CDNNs) in challenging inverse problems and high-dimensional regression tasks.

Deeply Virtual Compton Scattering (DVCS) offers a compelling setting in which to investigate these questions. As one of the most direct experimental probes of the three-dimensional structure of the nucleon, DVCS provides access to Generalized Parton Distributions (GPDs) through their convolution into Compton Form Factors (CFFs) \cite{Ji1997,Diehl03,Belitsky2002,Belitsky2010}. Extracting CFFs from experimental data is a central inverse problem in hadronic physics, complicated by the coherent interference of DVCS and Bethe–Heitler amplitudes, limited kinematic coverage, and strong correlations among fit parameters. Classical deep learning has already shown promise in addressing these challenges \cite{NNKumeriscki2011,Moutarde2019}, making DVCS an ideal testbed for assessing the potential added value of quantum approaches.

In this work, our primary focus is the development of practical tools for the systematic use of quantum deep neural networks and for determining, in advance, whether a CDNN or QDNN is better suited to a given extraction task. Central to this effort is the construction of a quantitative diagnostic we term the \emph{quantum qualifier}, designed to guide model selection based on intrinsic properties of the data. The application to experimental DVCS measurements from Jefferson Lab is presented as a culminating case study, illustrating how these methodological developments can inform deployment and regime selection for CFF extractions.

\section{\label{sec:QuantumDNNs}Quantum Deep Neural Networks}

Quantum deep neural networks (QDNNs) incorporate elements of quantum mechanics into learning architectures, providing a computational paradigm fundamentally different from that of classical deep neural networks (CDNNs). In a CDNN, each layer performs a transformation of the form \( h = \sigma(Wx + b) \), where \( W \) denotes the weight matrix, \( b \) is a bias vector, and \( \sigma \) represents a nonlinear activation function. By contrast, QDNNs implement quantum operations on qubits in place of these classical mappings. A complete quantum circuit is built from successive layers of parameterized unitary operators, interleaved with entangling gates that generate correlations among qubits. For a circuit composed of \( L \) layers, the resulting quantum state can be written as $|\psi_{\text{final}}\rangle = U_L U_{L-1} \cdots U_1 |0\rangle^{\otimes n}$.

A central component of any QDNN is the encoding of classical data into a quantum representation. Among the most widely used strategies is angle embedding, in which classical inputs are mapped to rotation angles of single-qubit gates. For an input feature \( x \), this encoding is realized through the operation $R_x(x)\ket{0} = e^{-i x X / 2} \ket{0}$. Following the quantum evolution, information is extracted through measurement. Classical outputs are obtained by evaluating expectation values of Hermitian observables, $\langle O \rangle = \bra{\psi} O \ket{\psi}$, which can subsequently be processed using conventional learning techniques. A typical QDNN architecture therefore comprises four key stages: data encoding, a sequence of parameterized unitary layers, entangling operations, and final measurement.

QDNNs are appealing because they exploit the exponential structure of quantum state space, enabling compact representations of complex transformations \cite{Biamonte2017}. A system of \(n\) qubits spans a Hilbert space of dimension \(2^n\), offering expressive power that can exceed the polynomial scaling typical of classical networks \cite{schuld2015introduction}. In addition, quantum entanglement and superposition provide natural mechanisms for modeling nontrivial correlations and parallel computational pathways \cite{havlicek2019supervised,schuld2018quantum}. Although a definitive demonstration of quantum advantage in deep learning remains open, accumulating evidence suggests that QDNNs may offer practical benefits for feature extraction and kernel-based learning in high-dimensional settings \cite{zoufal2019quantum,du2020expressive,abbas2021power}.

In the sections that follow, we examine systematic approaches for assessing the potential advantages of QDNNs over CDNNs in both classification and regression settings using well-defined performance metrics. Establishing robust evaluation tools is particularly important for applications in hadronic physics, where accurate and efficient information extraction is essential.

\subsection{Quantum Classification}

\begin{figure}[b!]
    \centering
    \subfloat[]{\includegraphics[width=0.495\textwidth]{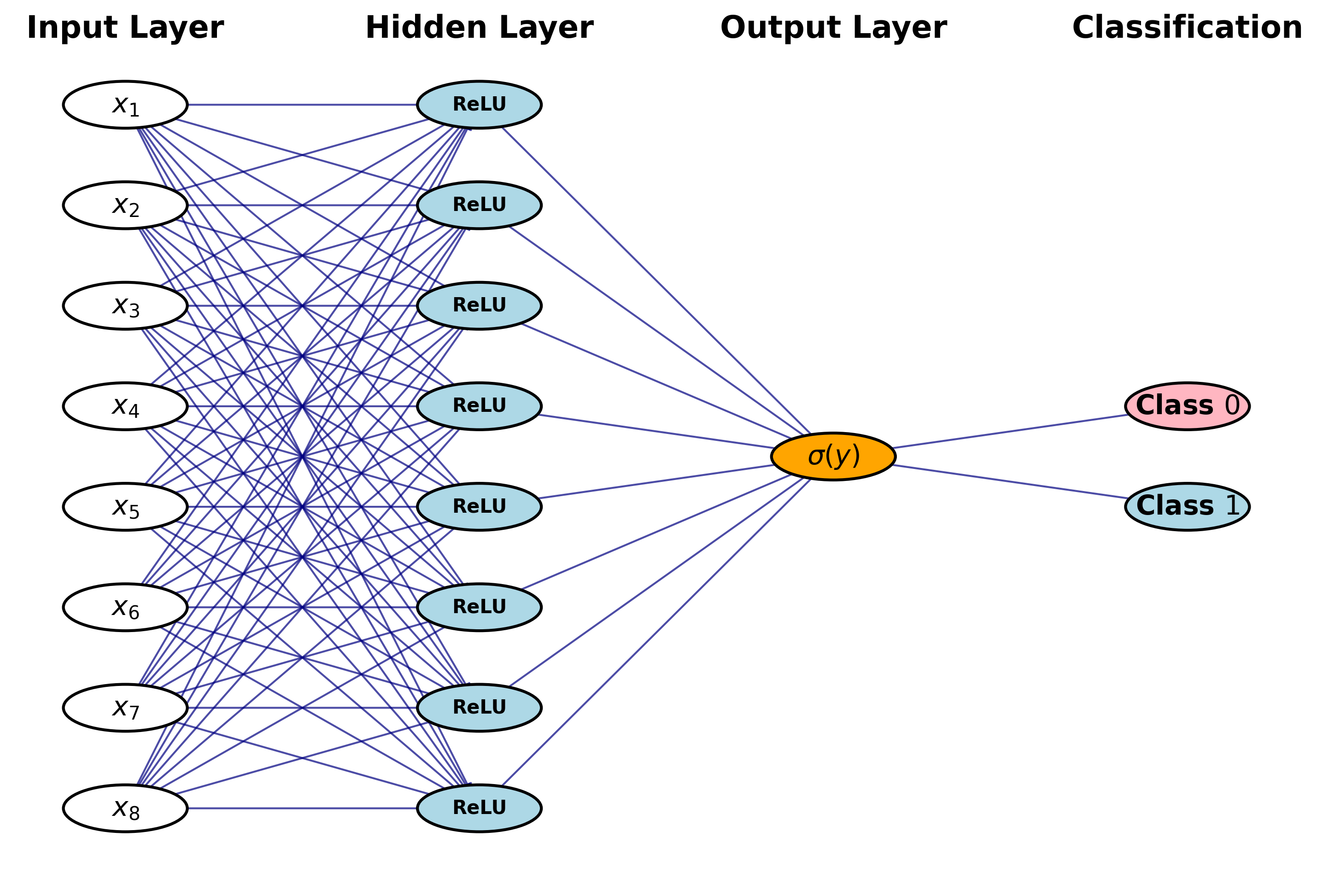}\label{fig:CDNN-arch}}
    \hfill
    \subfloat[]{\includegraphics[width=0.46\textwidth]{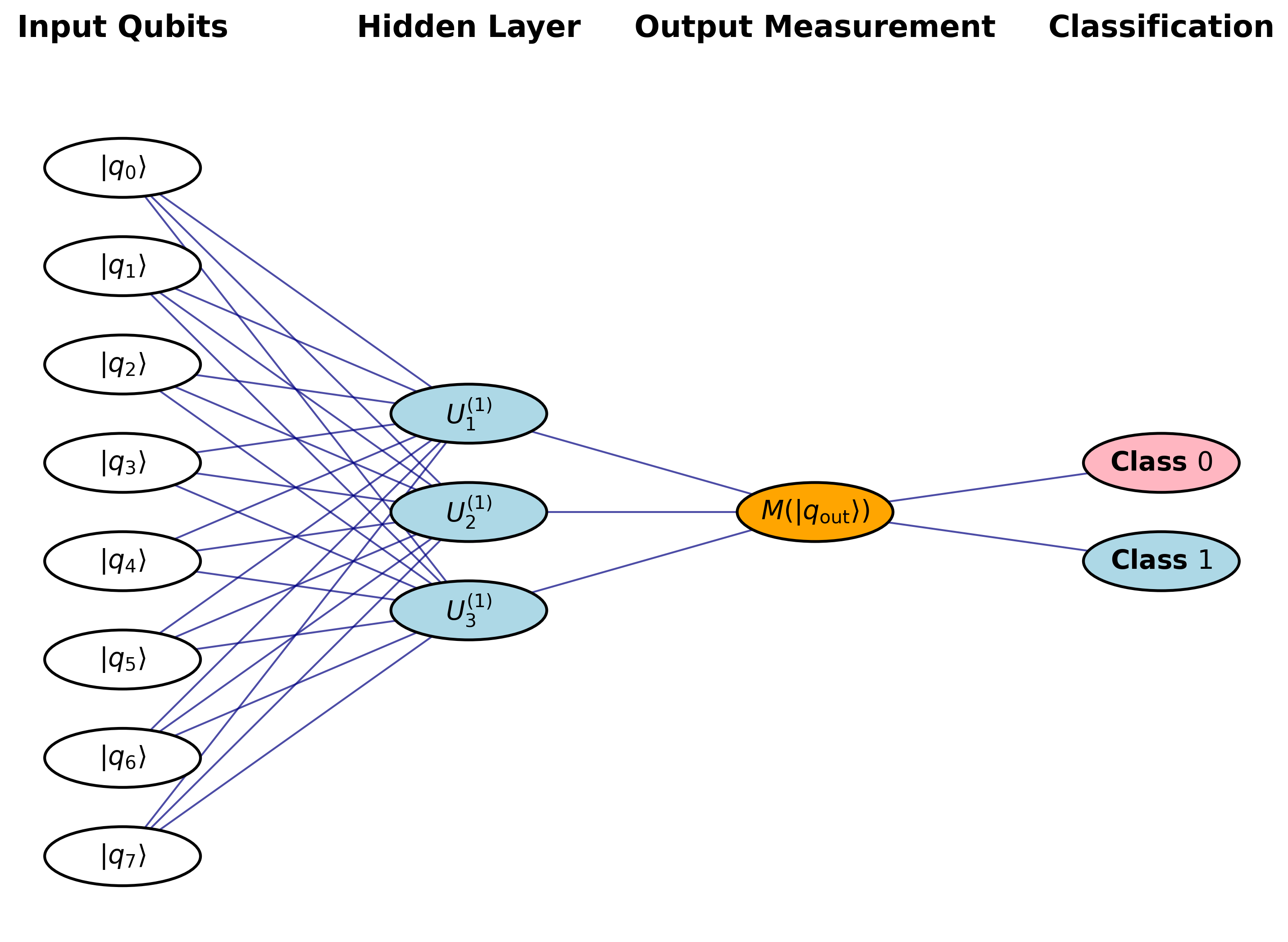}\label{fig:QDNN-arch}}
    \caption{\textbf{(a)}~Schematic of a CDNN for binary classification. Eight input features $x_i$ are passed through a hidden layer of 8 ReLU neurons, followed by a single sigmoid output neuron $\sigma(y)$ that returns the probability of belonging to Class~0 or Class~1. \textbf{(b)}~Schematic of a QDNN for classification. Eight input qubits $|q_i\rangle$ are processed by a layer of quantum perceptrons $U^{(1)}_i$. The final measurement $M(|q_{\mathrm{out}}\rangle)$ determines the class label.}
\end{figure}

To carry out a basic comparison between CDNN and QDNN classifiers, we construct a binary classification task using synthetically generated data from two distinct classes. Each data point is represented by a pair $\{X,\,y\}$, where $X \in \mathbb{R}^n$ denotes the vector of input features and $y \in \{0,1\}$ is the corresponding class label. In order to control the complexity of the learning problem, we consider two different types of datasets. The first, referred to as 1 function data, is generated by producing both classes from a single underlying function. The second, termed 3 function data, increases the level of complexity by generating the two classes from a combination of three distinct functions.

\begin{table*}[tb!]
    \centering
    \scriptsize
    \begin{tabular}{ccccc}
        Factor varied & Change & CDNN eff. & QDNN eff. & QDNN/CDNN ratio change \\
        \hline
        Number of training pairs & 500 pairs $\rightarrow$ 50 pairs & 0.6436 $\rightarrow$ 0.4716 & 0.8745 $\rightarrow$ 0.8116 & 27\% \\
        Data complexity & 1 func. data $\rightarrow$ 3 func. data & 0.8378 $\rightarrow$ 0.6290 & 0.9678 $\rightarrow$ 0.8704 & 20\% \\
        Number of input features & 8 feat. $\to$ 16 feat. & 0.6290 $\rightarrow$ 0.5327 & 0.8704 $\rightarrow$ 0.9359 & 27\% \\
        Noise & $0.2\sigma$ $\rightarrow$ $0.05\sigma$ & 0.7917 $\rightarrow$ 0.8378 & 0.8704 $\rightarrow$ 0.9678 & 5\%  \\
    \end{tabular}
    \caption{\label{tab:class-compar} Comparison of CDNN and QDNN classification efficiency for different data sets. Our default data set is 3 function data with 250 training pairs, 8 input features, and 0.05 noise; each row varies one of these factors while holding the rest at the default. The QDNN classification efficiency is higher than the CDNN classification efficiency for all datasets classified, indicating universal QDNN outperformance in this classification task. Fewer training pairs, higher data complexity, more input features, and lower noise lead to greater QDNN outperformance.}
\end{table*}
\begin{figure*}[b!]
    \centering
    \begin{subfigure}{0.45\textwidth}
        \centering
        \includegraphics[width=0.9\textwidth]{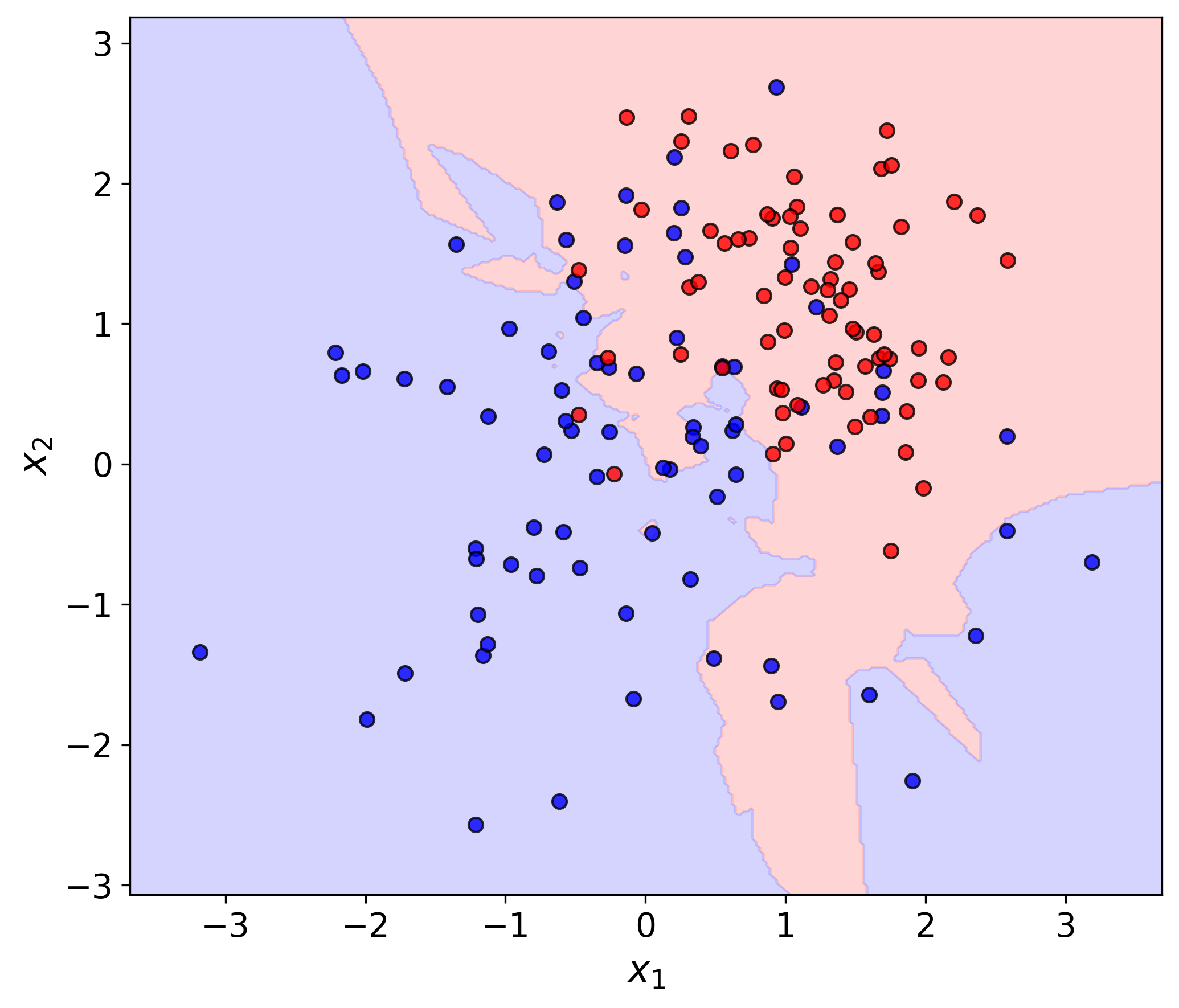}
        \caption{}
    \end{subfigure}
    \hfill
    \begin{subfigure}{0.45\textwidth}
        \centering
        \includegraphics[width=0.9\textwidth]{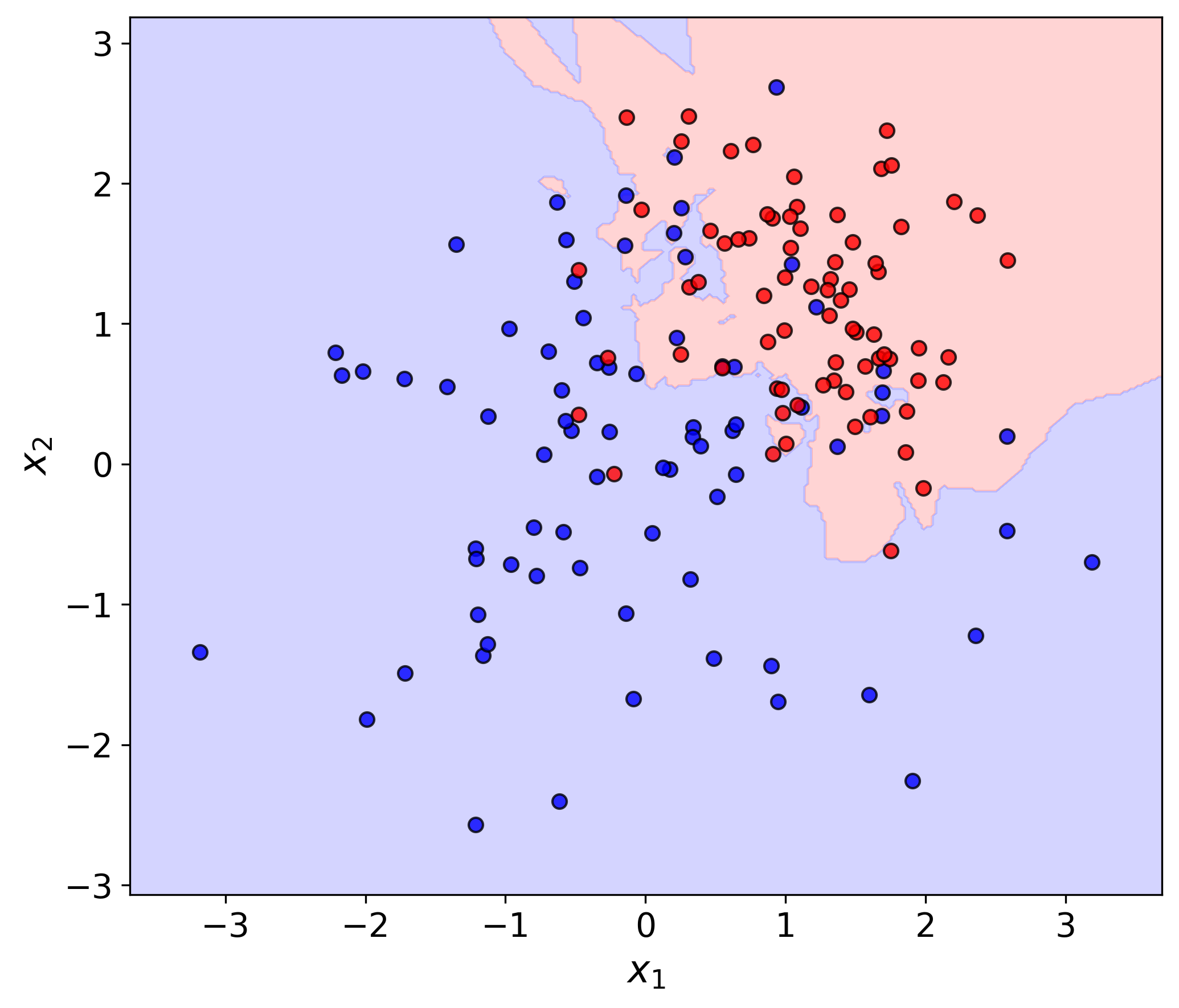}
        \caption{}
    \end{subfigure}
    \caption{Classification comparison between \textbf{(a)}~a CDNN and \textbf{(b)}~a QDNN using 1 function data with $0.2\sigma$ noise level and 8 input features. Two of the eight input features are plotted, with signal points in red and background points in blue. The QDNN outperforms the CDNN in classification accuracy.}
    \label{fig:class-compar}
\end{figure*}

Table~\ref{tab:class-compar} summarizes the response of our CDNN and QDNN classifiers to systematic variations in four key factors: the size of the training set, the complexity of the data, the number of input features, and the level of noise. In each study, one parameter is varied while the remaining three are held fixed. Across all datasets, the comparison of efficiencies reveals a consistent advantage of the QDNN over the CDNN in handling complex classification tasks. While noise contributes to this performance gap, the dominant factors driving QDNN outperformance are the number of training pairs, the intrinsic complexity of the data, and the dimensionality of the input space.

Although the relative advantage of QDNNs is most pronounced for complex datasets, the classification study shown in Fig.~\ref{fig:class-compar} demonstrates that a substantial performance gain persists even for a comparatively simple dataset. The corresponding confusion matrices \cite{duda2001pattern} for the CDNN ($\mathbf{C}^{\text{CDNN}}$) and QDNN ($\mathbf{C}^{\text{QDNN}}$) are
\begin{equation}
    \mathbf{C}^{\text{CDNN}} = \begin{bmatrix}
        70 & 24 \\ 4 & 52
    \end{bmatrix},\quad 
    \mathbf{C}^{\text{QDNN}} = \begin{bmatrix}
        65 & 6 \\ 9 & 70
    \end{bmatrix},
\end{equation}
corresponding to classification efficiencies of 0.8144 for the CDNN and 0.8998 for the QDNN.

Taken together, these results indicate a clear benefit to employing QDNNs for classification, consistent with earlier studies \cite{cong2019quantum,huang2022power,senokosov2024quantum,pesah2022quantum,alvarez2025benchmarking}, while highlighting the need for further work to assess the robustness of this advantage across broader datasets.

\subsection{Quantum Regression}
\label{sec:quantum-reg}

As an initial benchmark, we carried out a set of basic regression tests comparing the performance of a CDNN and a QDNN. For this purpose, we construct synthetic datasets by evaluating six target functions of increasing complexity on 100 uniformly spaced input values $x_i \in [-2,4]$. To emulate realistic conditions, we introduce noise into each dataset, with the standard deviation chosen as $\sigma = 0.1,\,0.25,$ and $1.0$ for every target function. To quantify the quality of a given neural network fit, we adopt a regression metric $M_{\text{reg}}$ defined as
\begin{equation}
    M_{\text{reg}} = \int_{x_{\text{min}}}^{x_{\text{max}}} \big|y_{\text{DNN}}(x) - y_{\text{true}}(x)\big|\dd{x}.
    \label{eq:mreg}
\end{equation}
To assess the relative performance of the two models, we introduce the quantum outperformance measure $\Xi$, defined for a given dataset and training duration as
\begin{equation}
    \Xi = M_{\text{reg}}^{\text{CDNN}}/M_{\text{reg}}^{\text{QDNN}} - 1.
    \label{eq:xi}
\end{equation}
By construction, $\Xi>0$ indicates superior performance of the QDNN, whereas $\Xi<0$ signifies that the CDNN provides a better fit. The absolute value of $\Xi$ reflects the degree of separation between the two models.

\begin{figure*}[tb!]
    \centering
    \begin{subfigure}{0.495\textwidth}
        \centering
        \includegraphics[width=\textwidth]{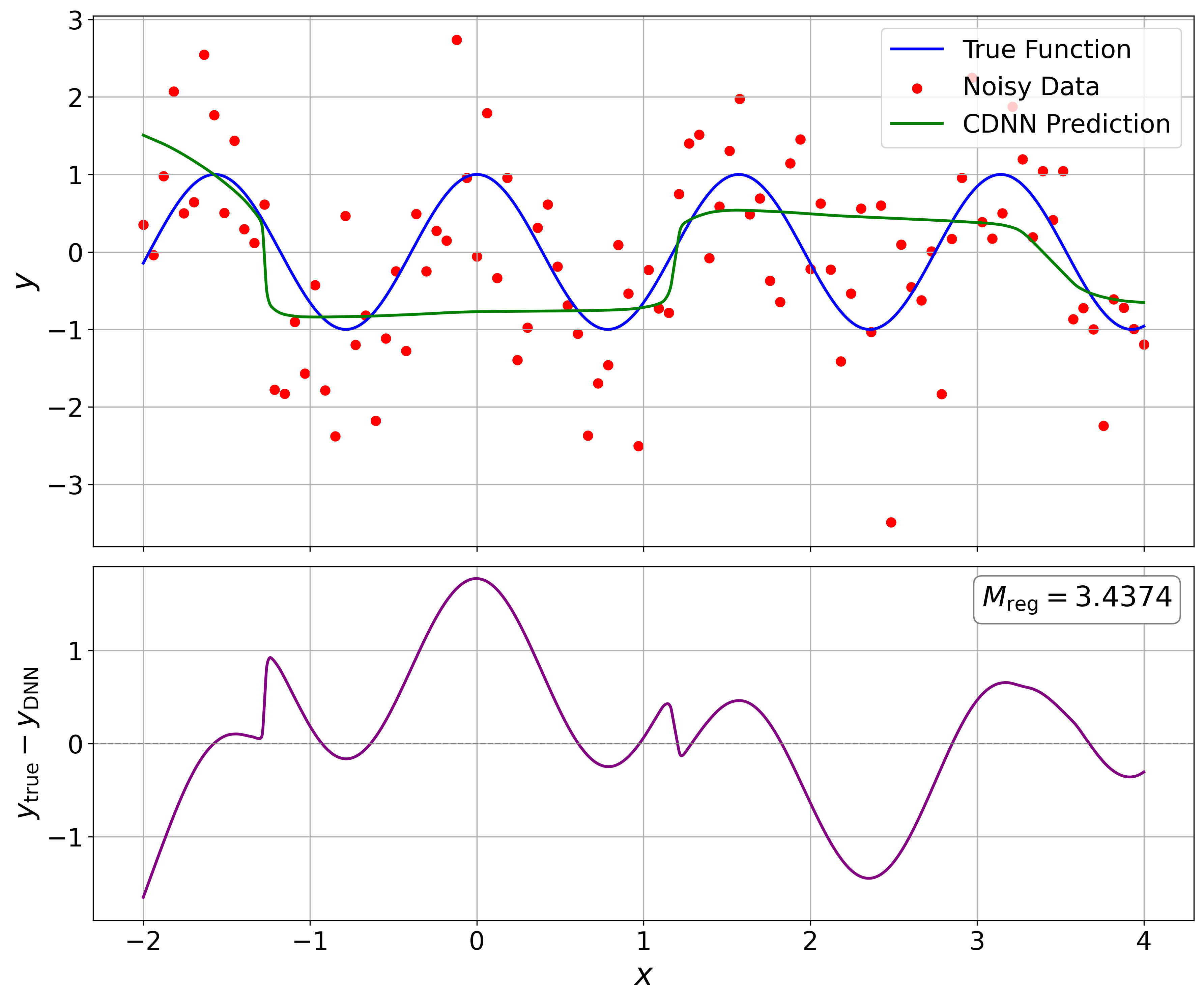}
        \caption{CDNN}
        \label{fig:c-reg-1}
    \end{subfigure}
    \begin{subfigure}{0.495\textwidth}
        \centering
        \includegraphics[width=\textwidth]{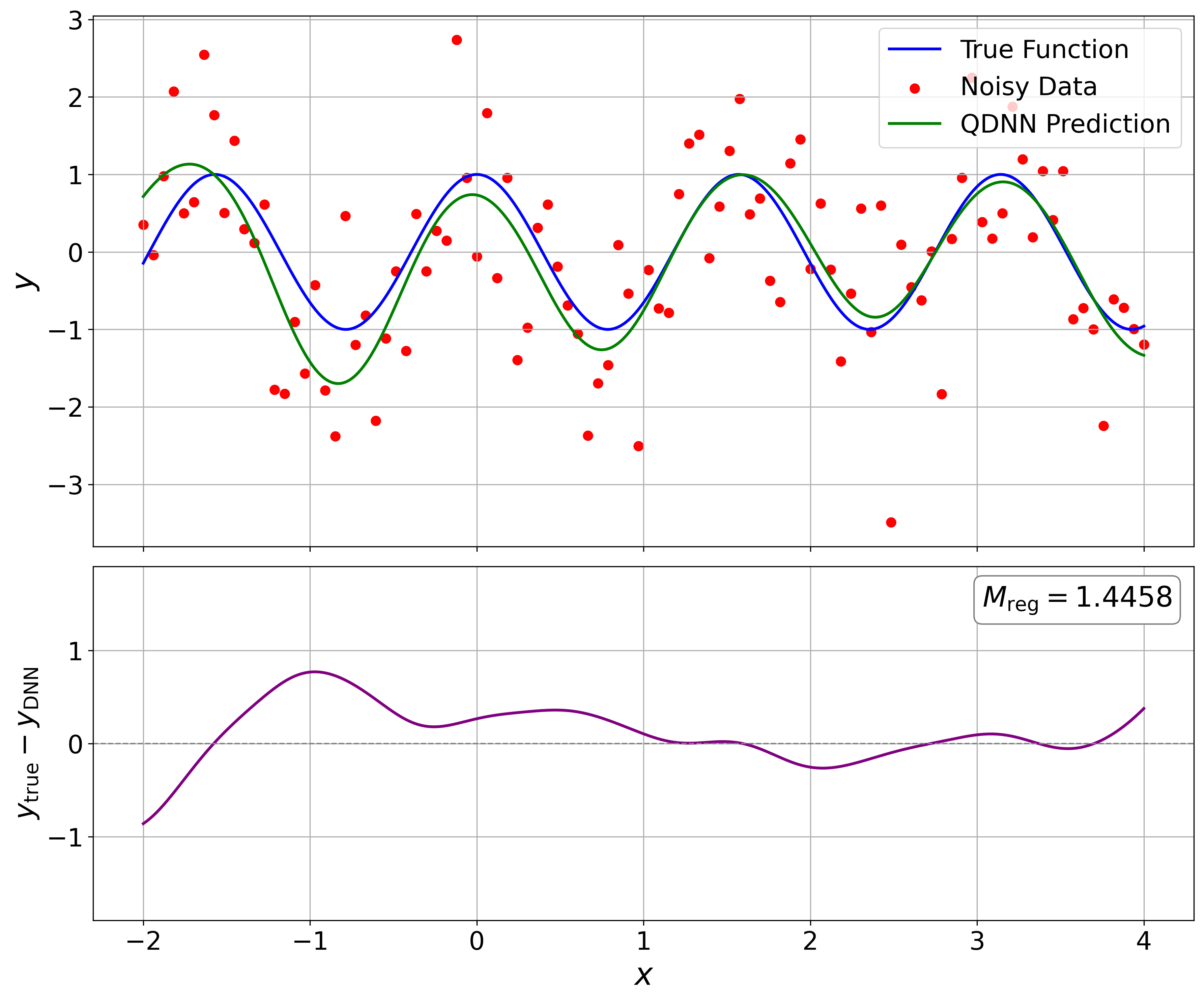}
        \caption{QDNN}
        \label{fig:q-reg-1}
    \end{subfigure}
    \caption{A function regression comparison between \textbf{(a)} a CDNN and \textbf{(b)} a QDNN using the target function $y = \cos 4x$ with 1$\sigma$ noise level and trained using 50 epochs.}
    \label{fig:reg-1}
\end{figure*}

A representative regression example is displayed in the upper panels of Fig.~\ref{fig:reg-1}, where the predictions of the CDNN and QDNN are compared against the true function. The corresponding difference between each prediction and the target function is shown in the lower panels. The area under the absolute difference curve, quantified by $M_{\text{reg}}$, is visibly smaller for the QDNN, demonstrating a reduced deviation relative to the CDNN. In this case, the QDNN achieves a clear advantage with an outperformance metric of $\Xi = 1.378 > 0$.

\section{Quantum Qualifiers}

Determining in advance whether a given function class or dataset is better suited for parameter extraction with a QDNN or a CDNN can offer a substantial methodological advantage. To enable a systematic assessment of this potential benefit, we introduce a collection of complexity and information-theoretic measures that together define a composite \textit{quantum qualifier}. Among the most informative of these indicators are the nonlinearity measure ($\mathfrak{N}$) \cite{montgomery2012,du2020expressive}, frequency complexity ($\Phi$) \cite{telgarsky2013dominant,Shen2021}, fractal dimension ($\mathfrak{D}$) \cite{buczkowski1998measurements,Verdon2023}, mutual information ($\mathfrak{M}$) \cite{Kraskov2004,Basilewitsch2025}, and Fourier transform complexity ($\mathfrak{F}$) \cite{Stoica2005,Chen2022}. Together, these metrics capture complementary structural and informational features of the target function or dataset and provide the foundation for constructing a composite quantum qualifier $\hat{\Xi}$ that estimates quantum outperformance $\Xi$ using data characteristics alone. We emphasize that the quantum qualifier is not intended as a proof of quantum advantage, but as a practical decision tool for model selection in data-driven analyses.

For a broad set of target functions evaluated at noise levels of $0.1\sigma$, $0.25\sigma$, and $1\sigma$, we compute these five characteristics and use Eq.~\eqref{eq:xi} to determine the corresponding quantum outperformance $\Xi$ for each test case. We then analyze the dependence of $\Xi$ on each characteristic at fixed numbers of training epochs. Linear regressions quantify these relationships through their slopes and associated $R^2$ values. To capture how the correlations evolve with training time, we model the regression slopes and coefficients of determination as polynomial functions of the epoch number $n$, using an exponential fit for the nonlinearity metric $\mathfrak{N}$. These fitted trends are used to construct a linear predictor for each observable at a given epoch, and the quantum qualifier $\hat{\Xi}$ is defined as a weighted sum of these predictors, with weights determined by the corresponding correlation strengths:
\begin{equation}
    \hat{\Xi} = e^{-\alpha n}\sum_{i=0}^{2}\beta_{1i}\, n^i X_1
    + \sum_{j=2}^{5}\sum_{i=0}^{4}\beta_{ji}\, n^i X_j.
    \label{eq:first-qq}
\end{equation}
The fitted values of the coefficients $\beta_{ji}$ are listed in Table~\ref{table:QQ-parms}. Equation~\eqref{eq:first-qq} indicates that mutual information $\mathfrak{M}$ exhibits the strongest positive correlation with quantum outperformance, whereas fractal dimension $\mathfrak{D}$ shows the largest and most pronounced negative correlation with $\Xi$.

\begin{table}[tb!]
\centering
\footnotesize
\setlength{\tabcolsep}{4pt}
\begin{tabular}{c c c c c c}
Metric $X_j$ & $\beta_{j0}$ & $\beta_{j1}$ & $\beta_{j2}$ & $\beta_{j3}$ & $\beta_{j4}$ \\
\hline
$\mathfrak{N}-0.25$ 
& $-1.17$ & $0.0163$ & $-2.65\times10^{-5}$ & --- & --- \\
$\Phi-24.5$ 
& $-0.0222$ & $2.82\times10^{-4}$ & $-1.40\times10^{-6}$ & $3.33\times10^{-9}$ & $-3.14\times10^{-12}$ \\
$\mathfrak{D}-0.95$ 
& $-1.10$ & $5.39\times10^{-3}$ & $-9.88\times10^{-6}$ & $7.96\times10^{-9}$ & $-4.06\times10^{-12}$ \\
$\mathfrak{M}+0.05$ 
& $0.548$ & $-6.57\times10^{-3}$ & $3.05\times10^{-5}$ & $-6.81\times10^{-8}$ & $6.05\times10^{-11}$ \\
$\mathfrak{F}-4999.5$ 
& $-5.97\times10^{-7}$ & $-1.48\times10^{-8}$ & $2.27\times10^{-11}$ & $-2.39\times10^{-14}$ & $2.03\times10^{-17}$ \\
\end{tabular}
\caption{\label{table:QQ-parms} Parameters for the double-summation expression of $\hat{\Xi}$ from Eq. \eqref{eq:first-qq}. The exponential prefactor is $\alpha=0.0101$.}
\end{table}

A dedicated quantum qualifier can be constructed for the deeply virtual Compton scattering (DVCS) process to assess whether QDNNs offer a tangible advantage for Compton form factor (CFF) extraction. To this end, we first define the DVCS quantum outperformance metric $\Xi_{DVCS}$, which the corresponding quantum qualifier $\hat{\Xi}_{DVCS}$ is designed to estimate. In direct analogy with the regression case, we introduce a goodness-of-fit measure
\begin{equation}
    M_{DVCS} = \int_{\phi_{\text{min}}}^{\phi_{\text{max}}}
    \big|F_{\text{DNN}}(\phi) - F_{\text{true}}(\phi)\big|\dd{\phi},
\end{equation}
where $F_{\text{DNN}}(\phi)$ is the cross section predicted from the DNN-extracted CFFs and $F_{\text{true}}(\phi)$ denotes the corresponding true cross section in the absence of experimental uncertainty. The quantum outperformance is then defined as
\begin{equation}
    \Xi_{DVCS} = M_{DVCS}^{\text{CDNN}}/M_{DVCS}^{\text{QDNN}} - 1.
\end{equation}
Following the same methodology used in the regression analysis, a DVCS-specific quantum qualifier $\hat{\Xi}_{DVCS}$ can be constructed \cite{le2025compton}. Notably, the procedure of testing datasets and constructing a quantum qualifier is broadly applicable: it can be implemented for any formalism, at any twist order, or for any choice of observables, with built-in sensitivity to experimental uncertainties and data sparsity. The development of such diagnostic metrics therefore holds significant promise for a wide range of applications in hadronic physics.

\begin{table*}[tb!]
\scriptsize
\begin{tabular}{lcccccc}
\multicolumn{1}{c}{Experiment} & Publication Year & $E_{beam}$ (GeV) & $Q^2$ (GeV$^2$) & $-t$ (GeV$^2$) & $x_B$ & Number of Points\\
\hline
Hall A E12-06-114 \cite{Georges2022} & 2022 & 4.487 - 10.992 & 2.71 - 8.51 & 0.204 - 1.373 & 0.363 - 0.617 & 1080\\
Hall A E07-007 \cite{Defurne2017} & 2017 & 3.355 - 5.55 & 1.49 - 2. & 0.177 - 0.363 & 0.356 - 0.361 & 404 \\
Hall A E00-110 \cite{Defurne2015} & 2015 & 5.75 & 1.82 - 2.37 & 0.171 - 0.372 & 0.336 - 0.401 & 468\\
Hall B e1-DVCS1 \cite{Jo2015} & 2015 & 5.75 & 1.11 - 3.77 & 0.11 - 0.45 & 0.126 - 0.475 & 1933\\ 
\end{tabular}
\caption{\label{tab:data_summary}Summary of the DVCS data from JLAB used in this analysis.}
\end{table*}

To ensure that these tests reflect realistic experimental conditions, the DVCS case study employs pseudodata constructed using the exact kinematics and uncertainty structure of DVCS measurements collected in Hall~A~\cite{Defurne2015, Defurne2017, Georges2022} and Hall~B~\cite{Jo2015} at Jefferson Lab (JLab). The underlying experimental dataset, summarized in Table~\ref{tab:data_summary}, contains both helicity-independent and helicity-dependent cross sections; for clarity and simplicity, our analysis is restricted to the helicity-independent (unpolarized) measurements.  The data are provided as finely binned fourfold differential cross sections in the kinematic variables \( Q^2 \), \( x_B \), \( t \), and \( \phi \), yielding a total of 3{,}885 data points. These measurements span both the 6~GeV and 12~GeV experimental programs at JLab. While the 12~GeV data extend to larger values of \( Q^2 \), reaching up to 8.4~GeV$^2$, the 6~GeV-era experiments populate the lower kinematic region. The bulk of the dataset originates from the e1-DVCS1 experiment in Hall~B~\cite{Jo2015}, which operated at a fixed beam energy of 5.75~GeV and covered a \( Q^2 \) range between 1.0 and 4.6~GeV$^2$.

\begin{figure*}[p!]
    \centering
    \begin{subfigure}{0.31\textwidth}
        \centering
        \includegraphics[width=\textwidth]{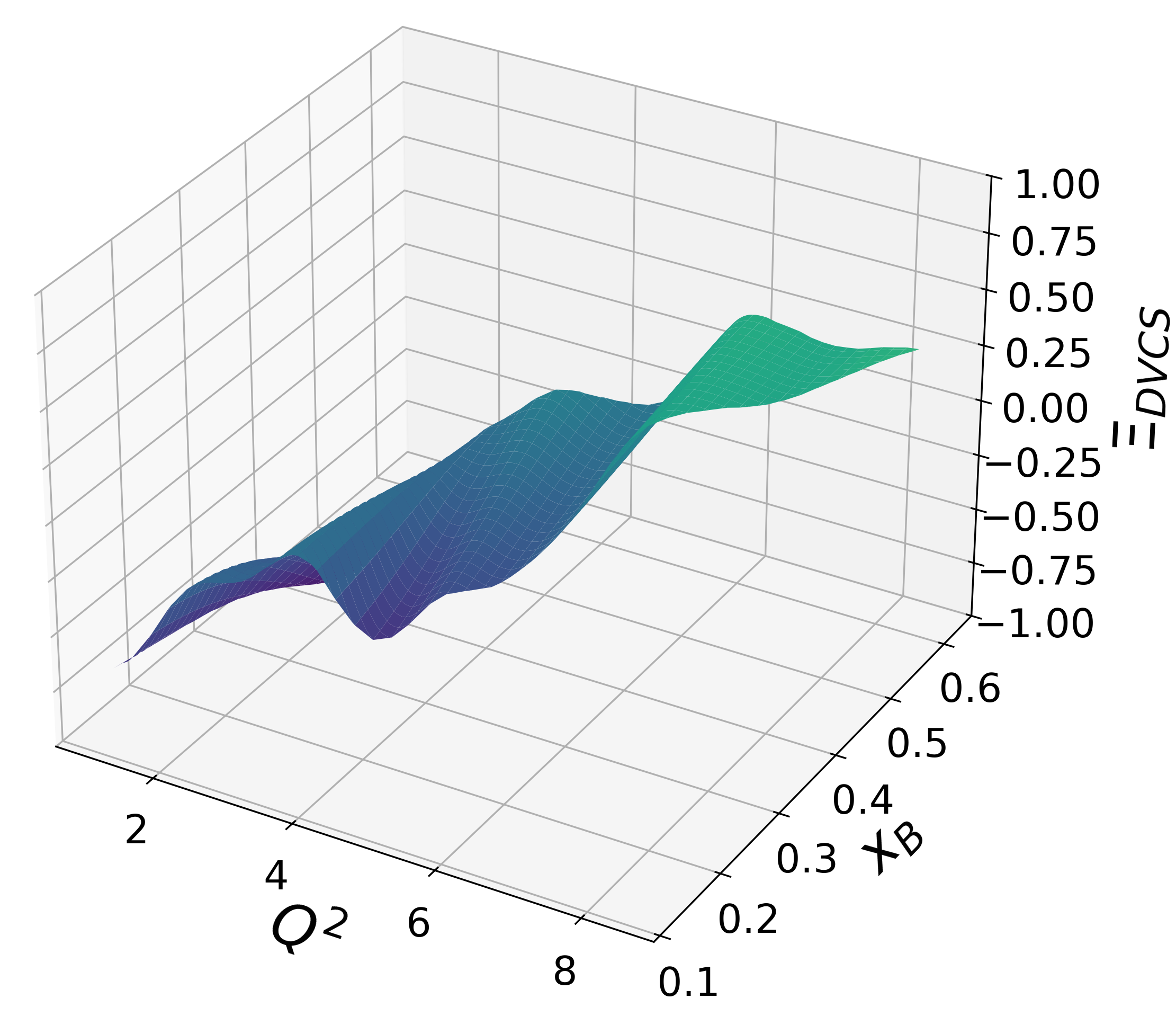}
        \caption{}
        \label{fig:xi_0.5sigma}
    \end{subfigure}
    \hfill
    \begin{subfigure}{0.31\textwidth}
        \centering
        \includegraphics[width=\textwidth]{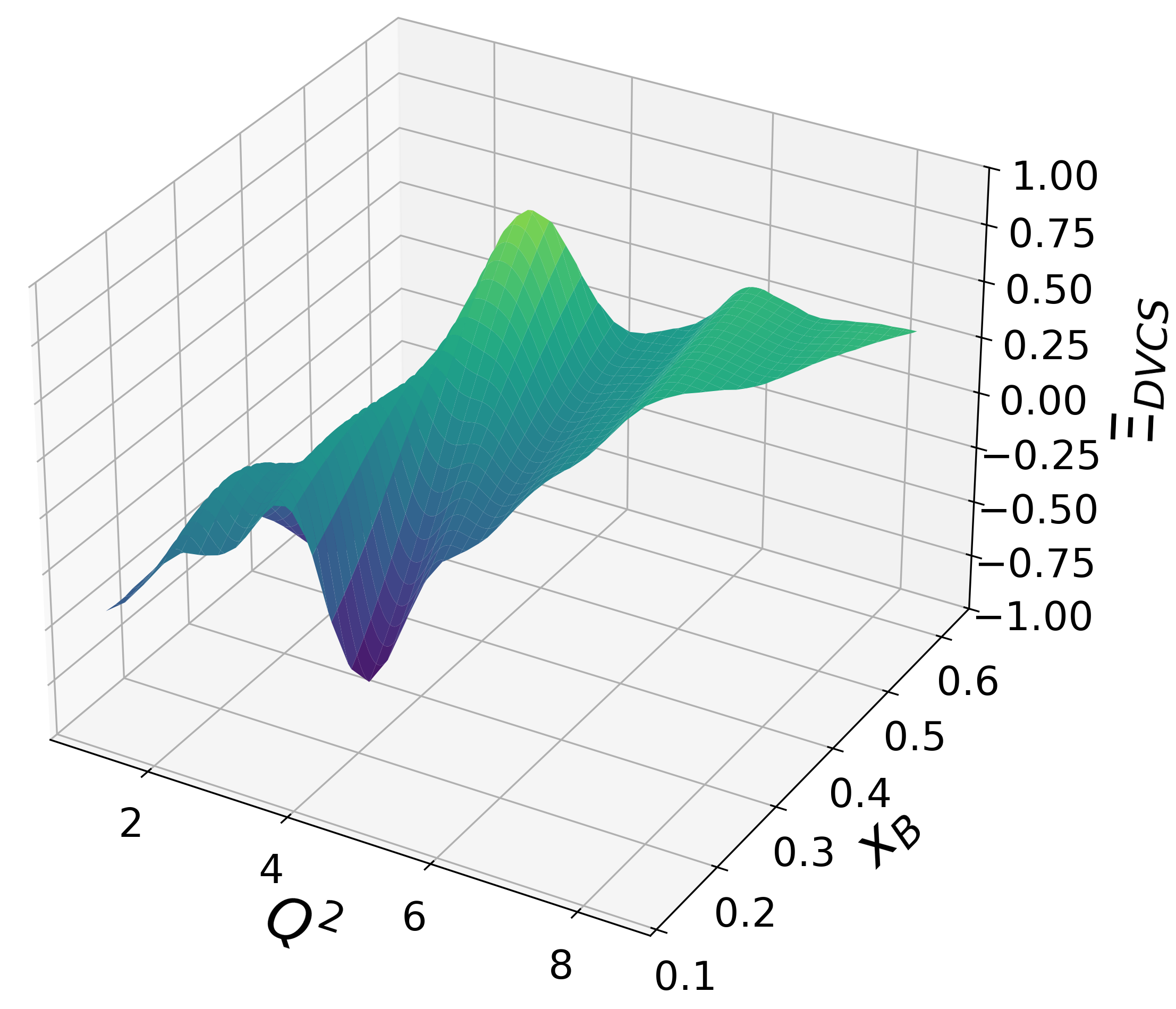}
        \caption{}
        \label{fig:xi_1sigma}
    \end{subfigure}
    \hfill
    \begin{subfigure}{0.36\textwidth}
        \centering
        \includegraphics[width=\textwidth]{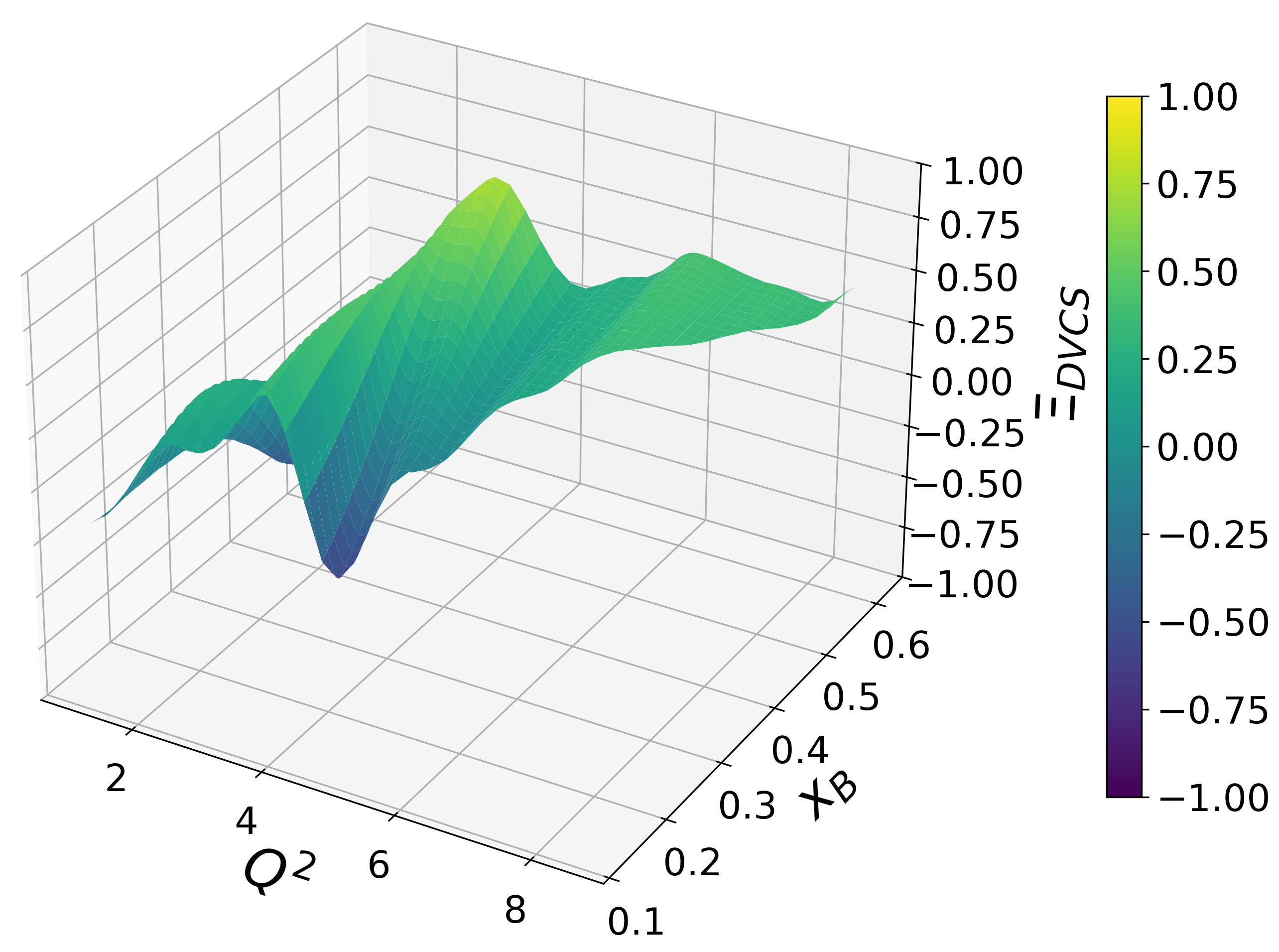}
        \caption{}
        \label{fig:xi_2sigma}
    \end{subfigure}
    \begin{subfigure}{0.31\textwidth}
        \centering
        \includegraphics[width=\textwidth]{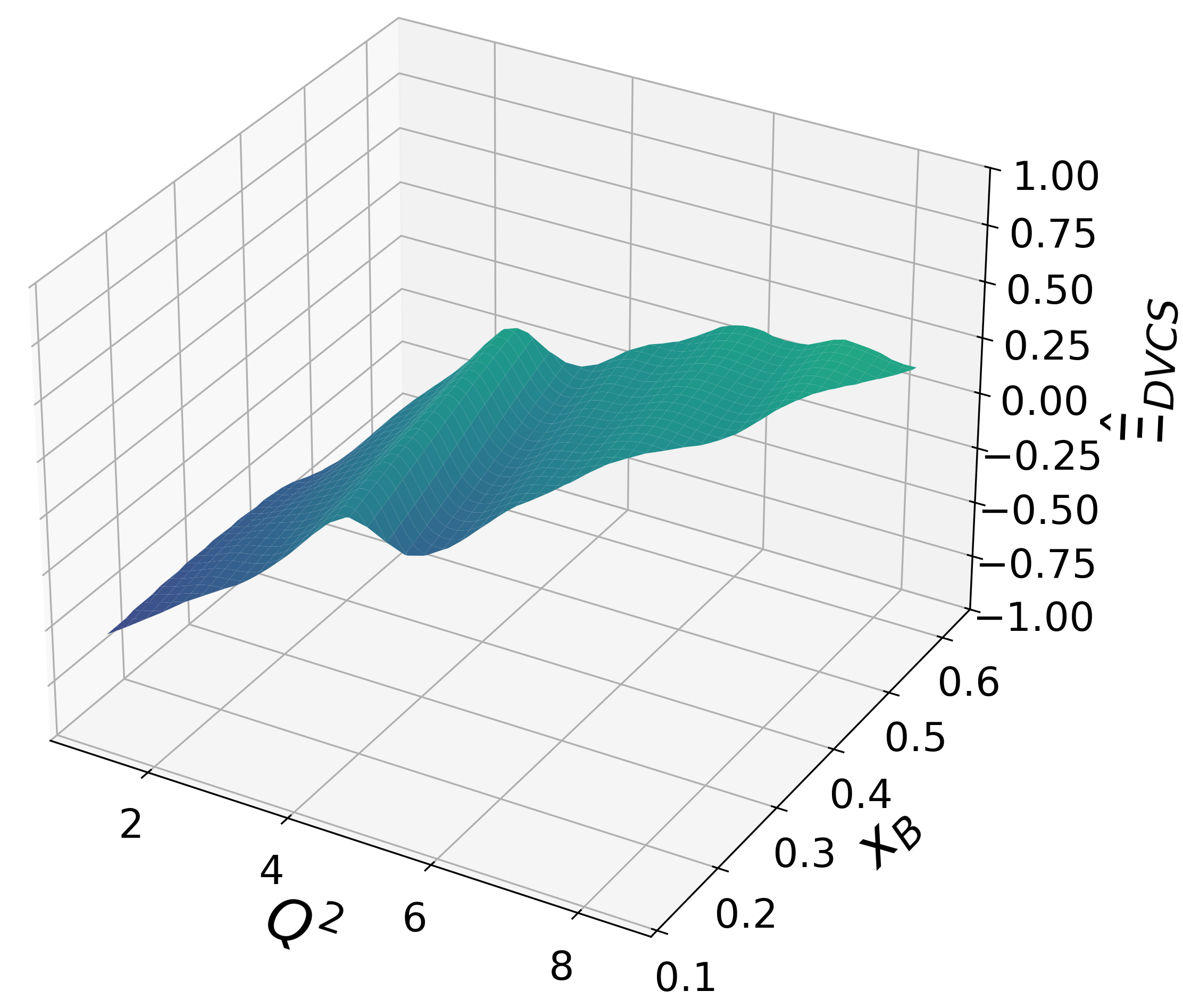}
        \caption{}
        \label{fig:xi_hat_0.5sigma}
    \end{subfigure}
    \hfill
    \begin{subfigure}{0.31\textwidth}
        \centering
        \includegraphics[width=\textwidth]{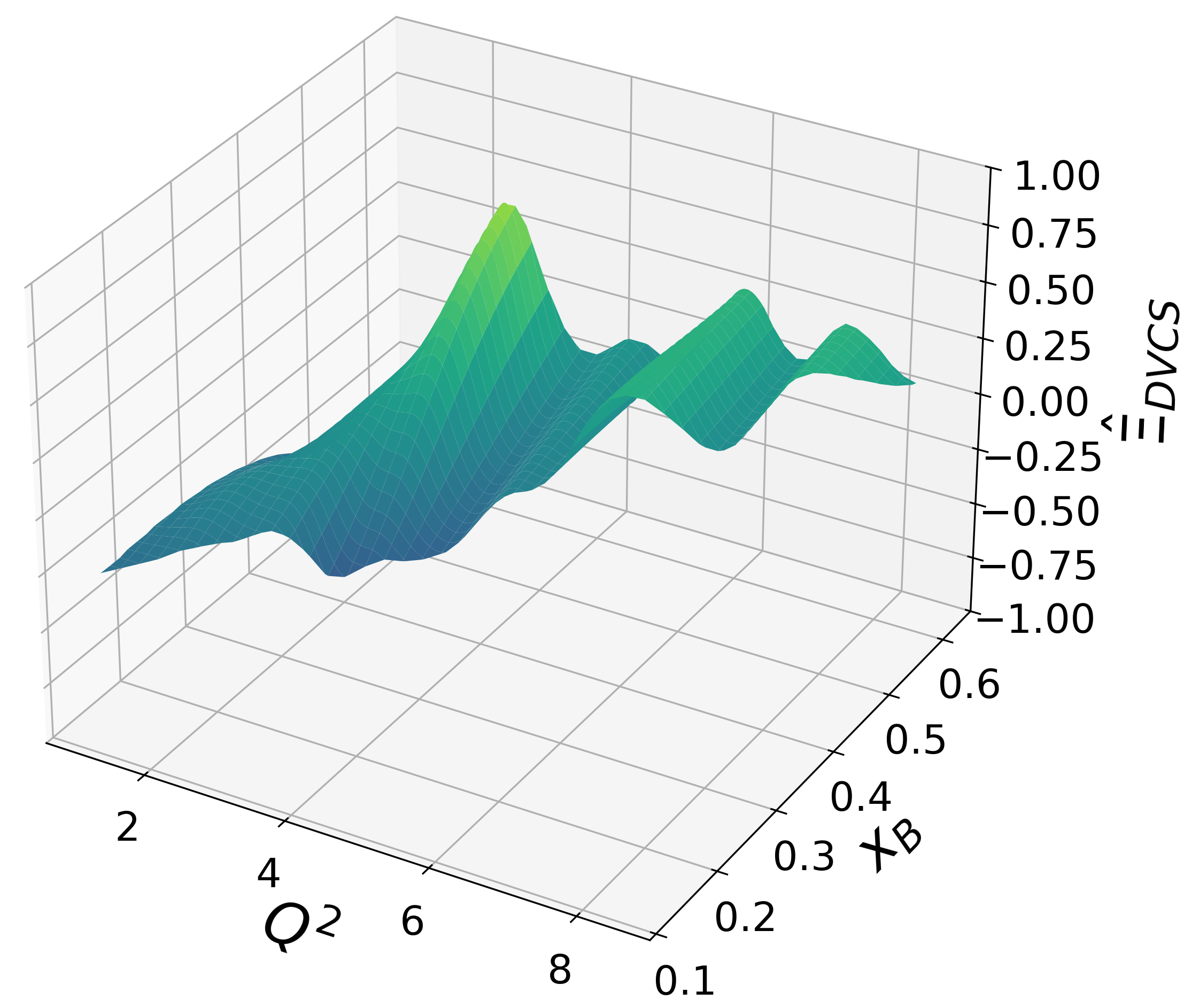}
        \caption{}
        \label{fig:xi_hat_1sigma}
    \end{subfigure}
    \hfill
    \begin{subfigure}{0.36\textwidth}
        \centering
        \includegraphics[width=\textwidth]{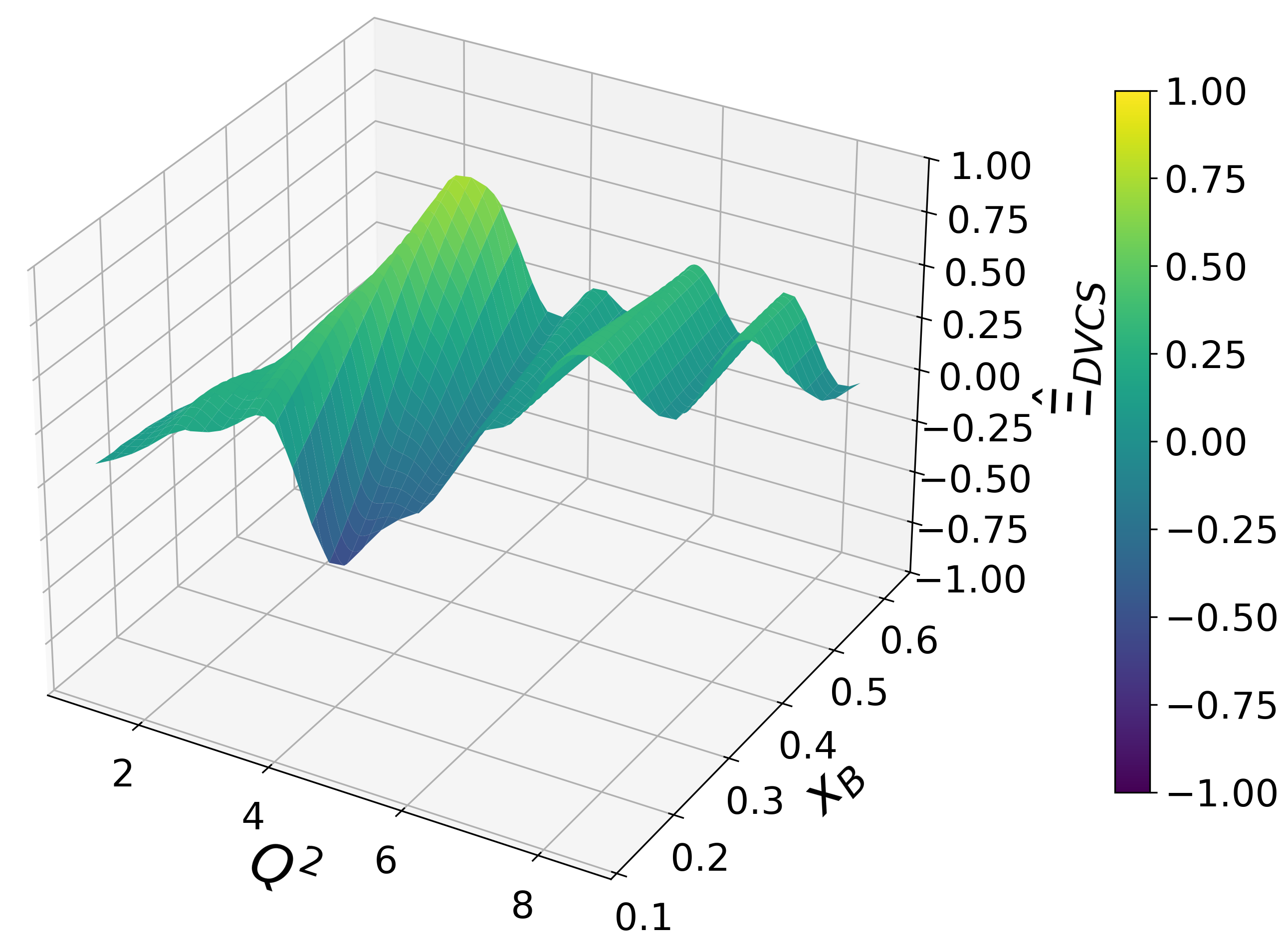}
        \caption{}
        \label{fig:xi_hat_2sigma}
    \end{subfigure}
    \caption{Comparison of the measured quantum outperformance $\Xi_{DVCS}$ and the DVCS quantum qualifier $\hat{\Xi}_{DVCS}$ across the experimental kinematic domain in $(Q^2,x_B)$ for three noise levels: \textbf{(a,d)}~$0.5\sigma$, \textbf{(b,e)}~$1\sigma$, and \textbf{(c,f)}~$2\sigma$. Surfaces are constructed from ensemble-averaged values using interpolation and smoothing within the convex hull of the data. The comparison illustrates how the quantum qualifier tracks the large-scale structure of the observed quantum outperformance and identifies kinematic regions where quantum models are expected to be advantageous.}
    \label{fig:xi-xihat-compar}
\end{figure*}

\begin{figure*}[p!]
    \centering
    \begin{subfigure}{0.3\textwidth}
        \centering
        \includegraphics[width=\textwidth]{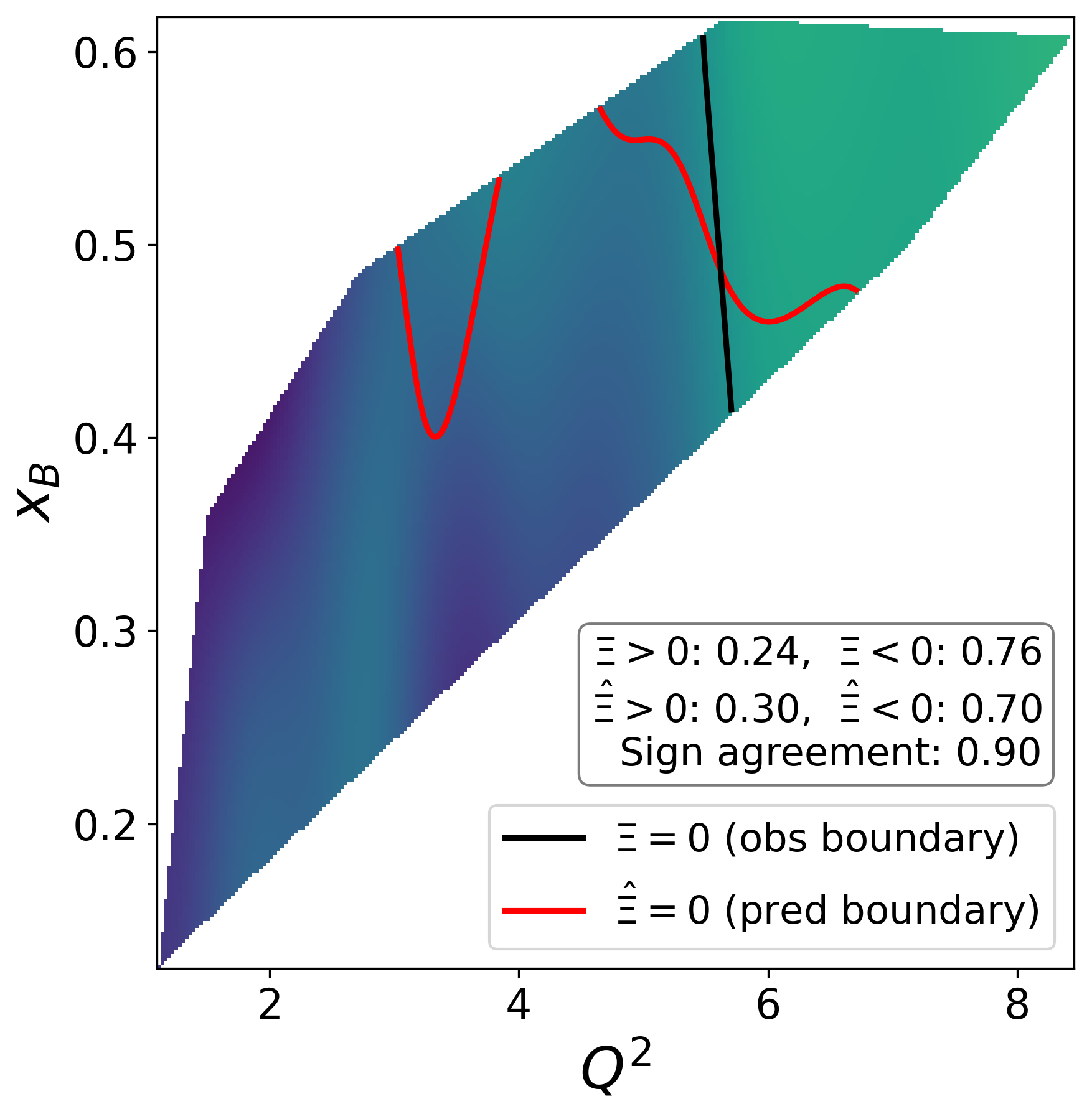}
        \caption{}
        \label{fig:bound_0.5sigma}
    \end{subfigure}
    \hfill
    \begin{subfigure}{0.3\textwidth}
        \centering
        \includegraphics[width=\textwidth]{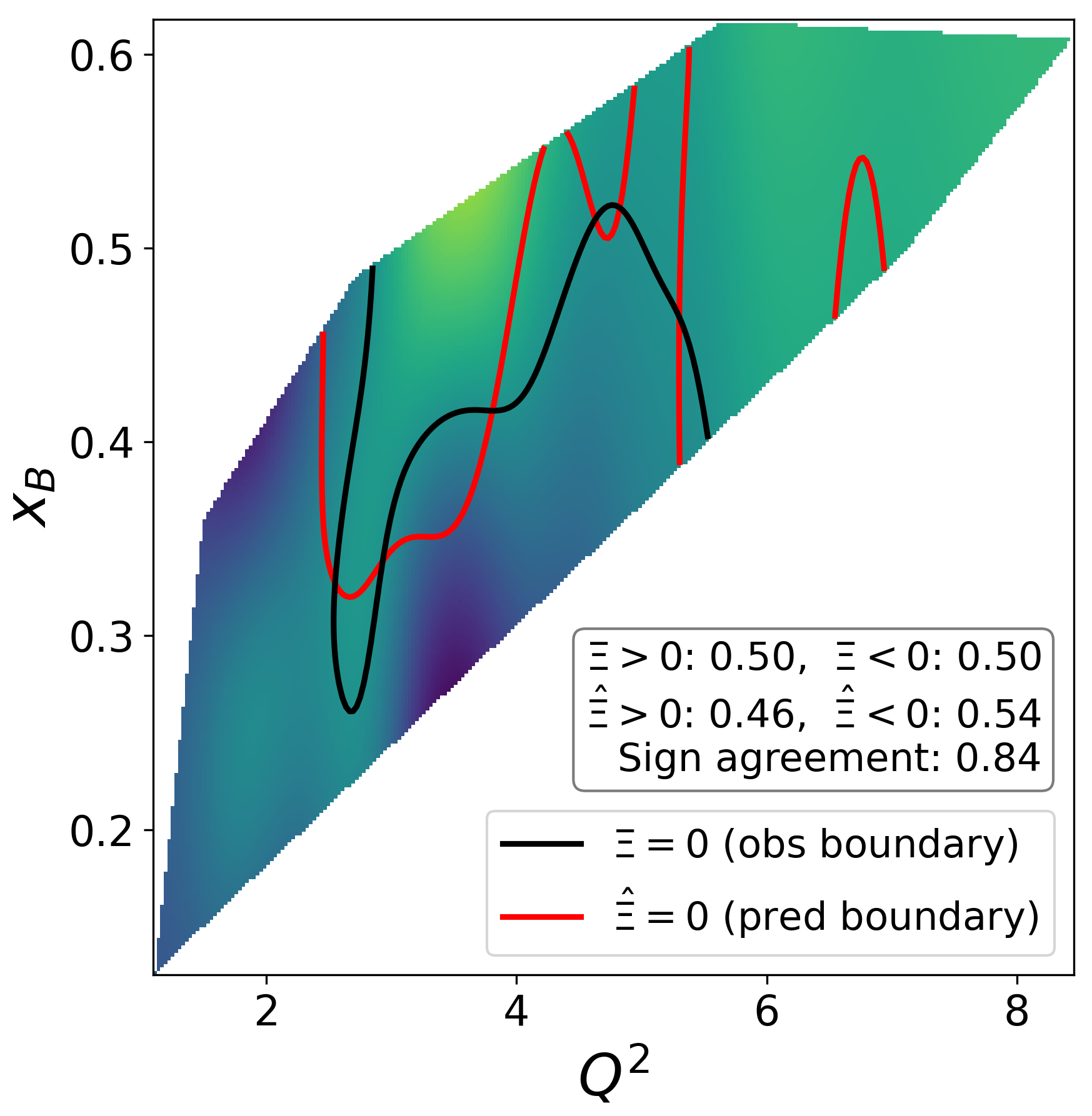}
        \caption{}
        \label{fig:bound_1sigma}
    \end{subfigure}
    \hfill
    \begin{subfigure}{0.38\textwidth}
        \centering
        \includegraphics[width=\textwidth]{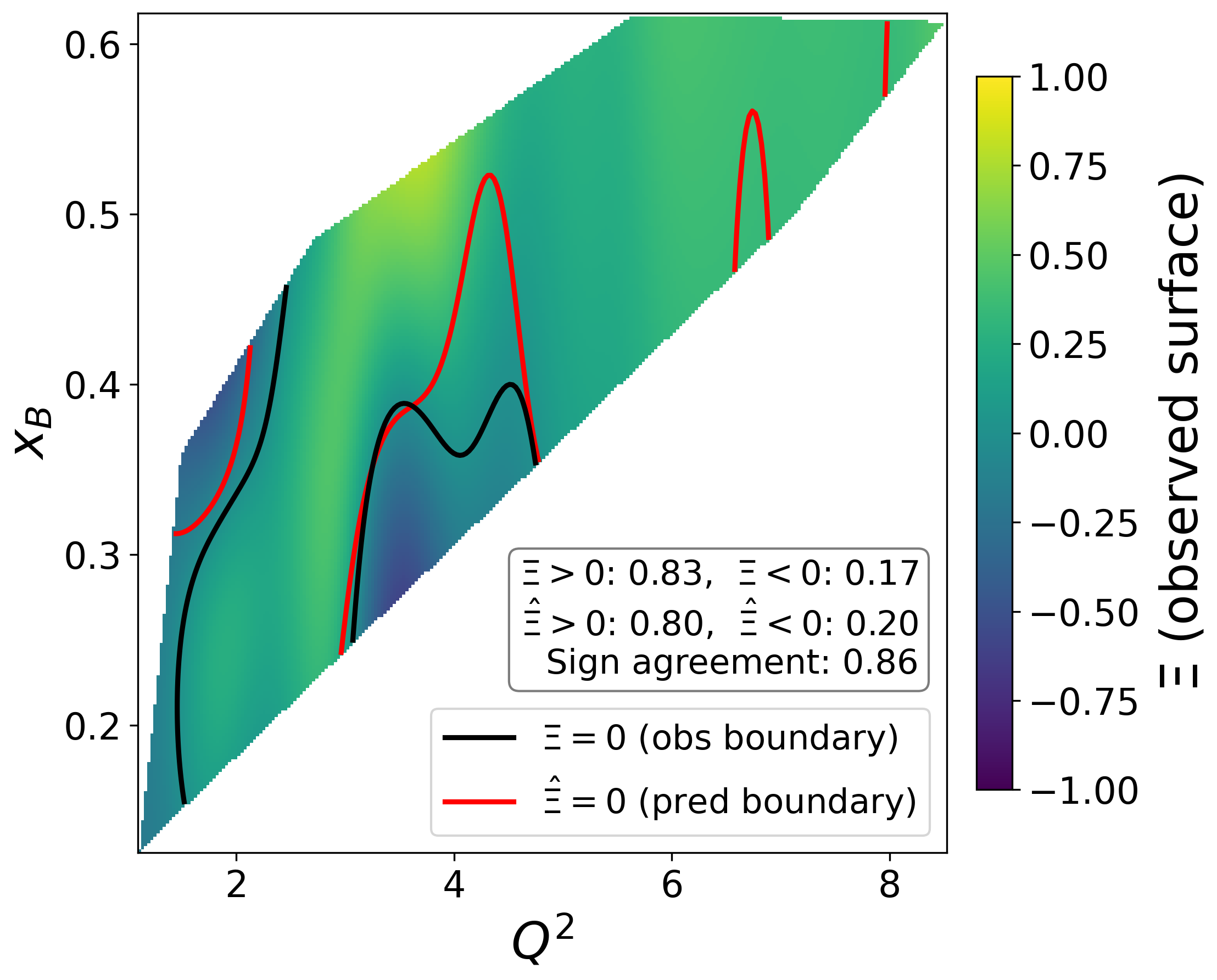}
        \caption{}
        \label{fig:bound_2sigma}
    \end{subfigure}
    \caption{Projections of the observed quantum outperformance surfaces $\Xi_{DVCS}(Q^2,x_B)$ for noise levels \textbf{(a)}~$0.5\sigma$, \textbf{(b)}~$1\sigma$, and \textbf{(c)}~$2\sigma$. The solid black curve denotes the empirical crossover boundary $\Xi_{DVCS} = 0$, while the solid red curve denotes the predicted crossover boundary $\hat{\Xi}_{DVCS} = 0$ obtained from the quantum qualifier. }
    \label{fig:boundaries}
\end{figure*}

Figure~\ref{fig:xi-xihat-compar} compares the measured quantum outperformance $\Xi_{DVCS}$ and the corresponding quantum qualifier $\hat{\Xi}_{DVCS}$ across the experimental kinematic domain for three representative noise levels, implemented through a controlled rescaling of the experimental uncertainties in the pseudodata construction. In all cases, the two surfaces exhibit strong qualitative agreement, with the quantum qualifier reproducing the dominant trends of the observed outperformance across $(Q^2,x_B)$. In particular, both $\Xi_{DVCS}$ and $\hat{\Xi}_{DVCS}$ display a systematic growth in both the magnitude and spatial extent of quantum advantage as the noise level increases. This consistency indicates that the quantum qualifier successfully captures the large-scale structure of quantum outperformance, supporting its role as a practical diagnostic for identifying kinematic regions where quantum models are most likely to be advantageous.

Figure~\ref{fig:boundaries} provides a two–dimensional projection of the outperformance surfaces shown in Figs.~\ref{fig:xi-xihat-compar}(a-c) onto the kinematic plane $(Q^2,x_B)$. Specifically, the colormap in each panel represents the observed outperformance surface $\Xi_{DVCS}(Q^2,x_B)$, evaluated within the convex hull of available kinematic points. Superimposed on this projection are the crossover boundaries defined by the zero level sets: the solid black curve indicates the empirical boundary $\Xi_{DVCS}(Q^2,x_B) = 0$ separating QDNN-favored ($\Xi_{DVCS}>0$) and CDNN-favored regions, while the solid red curve indicates the predicted boundary $\hat{\Xi}_{DVCS}(Q^2,x_B) = 0$ from the qualifier.

The inset statistics quantify the extent to which each model is favored across the accessible kinematic domain. For each noise level, we report the area fractions of the projected domain with $\Xi_{DVCS}>0$ and $\Xi_{DVCS}<0$, as well as the analogous fractions for $\hat{\Xi}_{DVCS}$. We also report a sign-agreement score, defined as the fraction of grid points for which $\sgn\Xi_{DVCS}(Q^2,x_B) = \sgn\hat{\Xi}_{DVCS}(Q^2,x_B)$, i.e., the fraction of the kinematic plane for which the qualifier predicts the correct side of the crossover. Across the three noise levels, the sign agreement remains high ($\sim 0.84$-$0.90$), indicating that the qualifier captures the dominant partitioning of the kinematic plane into QDNN- and CDNN-favored regimes. The area fractions provide a compact, geometry-based summary of the noise dependence already evident in the three-dimensional surfaces of Fig.~\ref{fig:xi-xihat-compar}. As the noise level increases from $0.5\sigma$ to $2\sigma$, the observed fraction $\text{Area}(\Xi_{DVCS}>0)$ increases from $0.24$ to $0.83$, which directly supports the conclusion that larger experimental uncertainty shifts the CFF extraction task toward regimes where the QDNN exhibits stronger outperformance.

Most importantly, the boundaries further reveal where in kinematic space the outperformance occurs. At low noise ($0.5\sigma$), the QDNN-favored region occupies a comparatively small subset of the domain, concentrated toward the upper portion of the accessible $(Q^2,x_B)$ range, while the remaining majority of the domain remains CDNN-favored. At intermediate noise ($1\sigma$), the boundaries traverse the interior of the domain and partition it into comparable QDNN- and CDNN-favored regions, reflecting a broad crossover. At high noise ($2\sigma$), the QDNN-favored region expands to cover most of the domain, with only localized CDNN-favored pockets remaining at lower $x_B$ and moderate $Q^2$. For intermediate to high noise levels, the most pronounced QDNN advantage is concentrated at large $x_B$ and low-to-intermediate $Q^2$, and as the noise level increases, this ``hot spot'' broadens into a larger QDNN-favored domain. In this way, the three-dimensional outperformance surfaces are converted into an explicit regime map in experimentally relevant kinematics, with the $\Xi_{DVCS} = 0$ contour serving as a practical decision boundary for model selection and the $\hat{\Xi}_{DVCS} = 0$ contour demonstrating that the qualifier reproduces this boundary with high accuracy.

We additionally probed the $t$-dependence by collapsing each experimental set to its mean $t$ and examining $\Xi_{DVCS}(t)$ and $\hat{\Xi}_{DVCS}(t)$. The resulting trends are consistent across noise levels: $\Xi_{DVCS}(t)$ is positive at large $-t$ (reaching $\sim 0.3$ near $t\simeq -1.3\text{ GeV}^2$ for $1\sigma$), crosses zero around $t\simeq -0.8\text{ GeV}^2$, and attains a minimum at low-to-intermediate $t$ (reaching $\sim-0.6$ near $t\simeq-0.2\text{ GeV}^2$ for $1\sigma$). To address whether the observed QDNN outperformance at large $-t$, $Q^2$, and $x_B$ could be an artifact of larger experimental uncertainties or sparser data there, we repeated the same analysis under two controls. First, we matched uncertainty by partitioning the sets into quantile bins of the per-set relative error $\overline{\epsilon}$ and recomputing the smoothed $\Xi_{DVCS}$ trends within each $\overline{\epsilon}$-bin, so each curve compares only sets with similar experimental precision. Second, we matched sampling density by repeating the analysis after restricting to subsets with the highest set counts, thereby removing the sparsest regions from the fit. In both cases, all qualitative trends established in the full-sample analysis are preserved, indicating that the observed QDNN-favored behavior is not solely a consequence of increased error bars or data sparsity in extreme kinematics.

\section{Conclusion}

In this work, we developed a systematic framework for assessing the applicability of quantum deep neural networks in data-driven problems relevant to hadronic physics. Rather than focusing solely on performance comparisons, our primary contribution is the development of a systematic framework for constructing practical diagnostic tools—most notably the \emph{quantum qualifier}—that assess, in advance, whether a CDNN or QDNN is better suited to a given extraction task. Through controlled classification and regression benchmarks, we demonstrated how QDNN performance depends sensitively on data complexity, noise, and dimensionality, and showed that these dependencies can be quantified using complexity- and information-theoretic measures. This approach shifts the emphasis from post hoc model comparison to principled model selection, providing a scalable methodology for guiding the use of quantum machine-learning architectures.

As a culminating application, we applied this framework to DVCS and showed that the qualifier not only tracks the overall noise dependence of quantum outperformance, but also reproduces the kinematic partitioning of the $(Q^2,x_B)$ plane into QDNN- and CDNN-favored regions. In particular, increasing noise systematically expands the QDNN-favored domain, and across noise levels the QDNN advantage is generally strongest toward large $Q^2$, $x_B$, and $-t$, a trend that persists under uncertainty-matched and density-controlled checks, indicating it is not solely an artifact of larger experimental errors or sparser coverage in those regions. This establishes the qualifier as a practical, experimentally interpretable decision boundary for model selection. More broadly, the tools developed here are not limited to DVCS or to CFF extraction, but can be adapted to a wide range of observables, theoretical formalisms, and experimental settings. As quantum computing hardware and hybrid quantum--classical workflows continue to mature, such data-informed strategies will be essential for integrating quantum machine learning into precision studies of hadronic structure.

\bibliographystyle{apsrev4-2}
\bibliography{refs}

\end{document}